# A principled analysis of merging operations in possibilistic logic

Salem Benferhat, Didier Dubois, Souhila Kaci and Henri Prade
Institut de Recherche en Informatique de Toulouse (I.R.I.T.)-C.N.R.S.
Université Paul Sabatier, 118 route de Narbonne 31062 TOULOUSE Cedex 4, FRANCE
E-mail:{benferhat, dubois, kaci, prade}@irit.fr

## Abstract

Possibilistic logic offers a qualitative framework for representing pieces of information associated with levels of uncertainty or priority. The fusion of multiple sources information is discussed in this setting. Different classes of merging operators are considered including conjunctive, disjunctive, reinforcement, adaptive and averaging operators. Then we propose to analyse these classes in terms of postulates. This is done by first extending the postulates for merging classical bases to the case where priorities are available.

## 1 Introduction

Possibilistic logic (e.g. [8]) offers a framework for reasoning with classical logic formulas associated with weights belonging to a totally ordered scale. Weights, which technically speaking are lower bounds of necessity measures, can either represent the certainty with which the associated formula is held for true, or the expression of a preference under the form of a level of priority. In this case the formula encodes a goal (rather than a piece of knowledge) which has to be considered. The fusion of information expressed in a logical form has raised an increasing interest in the recent past years [1, 6, 10, 11, 13, 14]. Indeed this problem naturally occurs when handling multiple sources of information, and trying to extract the common, conflict-free part of the information, or when trying to fuse the goals expressed by several agents. Clearly possibilistic logic, which offers a representation framework more expressive than the one of classical logic, by allowing for an explicit stratification of the sets of formulas, is well-suited for handling levels of certainty or priority in the fusion process. In recent works [3, 5], the authors have on the one hand provided a possibilistic syntactic counterpart of combination operations defined on possibility distributions defined on sets of interpretations. On the other hand, taking advantage of the fact that a classical logic formula can be always associated with a stratified set of formulas (using Hamming distance as suggested by Dalal [7]) which reflects partial levels of satisfaction of the initial formula, the authors have shown the agreement of the possibilistic logic-based approach with the recent proposals on fusion in the classical logic setting.

In this paper we make a step further by $i$) distinguishing between different classes of combination operations capable of coping with redundancy, or with drowning effects of "inconsistency-free" formulas [2] encountered in case of conflicts when weights are just combined by a simple operator like $min$, and $ii$) by analysing these classes firstly in terms of information sets that each class retains, and secondly in terms of postulates which are natural extensions of those recently proposed in the classical framework [10, 11, 12]. After briefly recalling the necessary background on possibilistic logic in Section 2, general classes of combination operators are introduced and studied in Section 3. The handling of the global reliability of the sources or of priorities between agents is also briefly considered in this section. A discussion with respect to postulates is presented in Sections 4 and 5.

## 2 Possibilistic logic and fusion

This section recalls some basic notions of possibilistic logic. See [8] for more details. Let $\mathcal{L}$ be a finite propositionnal language. $\vdash$ denotes the classical consequence relation and $\Omega$ is the set of classical interpretations.

### 2.1 Possibility distributions

At the semantic level, possibilistic logic is based on the notion of a possibility distribution, denoted by $\pi$, which is a mapping from $\Omega$ to $[0,1]$ representing the available information. $\pi(\omega)$ represents the degree of compatibility of the interpretation $\omega$ with the available beliefs about the real world if we are representing uncertain pieces of knowledge (or the degree of satisfaction of reaching state $\omega$ if we are modelling preferences). By convention, $\pi(\omega) = 1$ means that it is totally possible for $\omega$ to be the real world (or that $\omega$ is fully satisfactory), $1 > \pi(\omega) > 0$ means that $\omega$ is only somewhat possible (or satisfactory), while $\pi(\omega) = 0$ means that $\omega$ is certainly not the real world (or not satisfactory at all). Associated with a possibility distribution $\pi$ is the necessity degree of any formula $\phi$: $N(\phi) = 1 - \Pi(\neg \phi)$ which evaluates to what extent $\phi$ is entailed by the available beliefs, and defined from the consistency degree of a formula $\phi$ w.r.t. the available information, $\Pi(\phi) = max\{\pi(\omega) : \omega \in [\phi]\}$, where $[\phi]$ denotes the set of all the models of $\phi$.

In the rest of the paper, $a, b, c, ...$ reflect the possibility degrees of the interpretations.

## 2.2 Possibilistic logic bases

At the syntactic level, uncertain information is represented by means of *a possibilistic knowledge base* which is a set of weighted formulas $B = \{(\phi_i, \alpha_i) : i = 1, n\}$ where $\phi_i$ is a classical formula and $\alpha_i$ belongs to a totally ordered scale such as $[0,1]$. $(\phi_i, \alpha_i)$ means that the certainty degree of $\phi_i$ is at least equal to $\alpha_i$ ($N(\phi_i) \geq \alpha_i$). We denote by $B^*$ the classical base associated with $B$ obtained by forgetting the weights. A possibilistic base $B$ is consistent iff its classical base $B^*$ is consistent.

In the following, $\alpha, \beta, \gamma, ...$ reflect the necessity degrees associated with formulas.

Given $B$, we can generate a unique possibility distribution, denoted by $\pi_B$, such that all the interpretations satisfying all the beliefs in $B$ will have the highest possibility degree, namely 1, and the other interpretations will be ranked w.r.t. the highest belief that they falsify, namely we get [8]:

**Definition 1** $\forall \omega \in \Omega$,

$$\pi_B(\omega) = \begin{cases} 1 & if\ \forall (\phi_i, \alpha_i) \in B, \omega \in [\phi_i] \\ 1 - max\{\alpha_i : \omega \notin [\phi_i]\} & otherwise. \end{cases}$$

Further definitions used in the paper are now given:

**Definition 2** *Let $B$ be a possibilistic knowledge base, and $\alpha \in [0,1]$. We call the $\alpha$-cut (resp. strict $\alpha$-cut) of $B$, denoted by $B_{\geq \alpha}$ (resp. $B_{> \alpha}$), the set of classical formulas in $B$ having a certainty degree at least equal to $\alpha$ (resp. strictly greater than $\alpha$).*

**Definition 3** *$B$ and $B'$ are said to be equivalent, denoted by $B \equiv_s B'$, iff $\forall \alpha \in [0,1]$, $B_{\geq \alpha} \equiv B'_{\geq \alpha}$, where $\equiv$ is the classical equivalence.*

$Inc(B) = max\{\alpha_i : B_{\geq \alpha_i}\ is\ inconsistent\}$ denotes the inconsistency degree of $B$. When $B$ is consistent, we have $Inc(B) = 0$.

Subsumption can now be defined:

**Definition 4** *Let $(\phi, \alpha)$ be a belief in $B$. Then, $(\phi, \alpha)$ is said to be subsumed by $B$ if $(B - \{(\phi, \alpha)\})_{\geq \alpha} \vdash \phi$. $(\phi, \alpha)$ is said to be strictly subsumed by $B$ if $B_{> \alpha} \vdash \phi$.*

It can be checked that if $(\phi, \alpha)$ is subsumed, then $B$ and $B' = B - \{(\phi, \alpha)\}$ are equivalent [8].

Lastly, weights are propagated in the inference process:

**Definition 5** *A possibilistic formula $(\phi, \alpha)$, with $\alpha > Inc(B)$, is said to be a consequence of $B$, denoted by $B \vdash_\pi (\phi, \alpha)$, iff $B_{\geq \alpha} \vdash \phi$.*

## 2.3 Syntactic fusion

We first recall a general result underlying the fusion process in possibilistic logic [5].

Let $B_1$, $B_2$ be two possibilistic bases, and $\pi_1$ and $\pi_2$ be their associated possibility distributions. Let $\oplus$ be a two place function whose domain is $[0,1] \times [0,1]$ (to be used for aggregating $\pi_1(\omega)$ and $\pi_2(\omega)$). The only requirements for $\oplus$ are the following properties:

$i.\ 1 \oplus 1 = 1$,

$ii.$ If $a \geq c$, $b \geq d$ then $a \oplus b \geq c \oplus d$ (monotonicity).

The first one acknowledges the fact that if two sources agree that $\omega$ is fully possible (or satisfactory), then the result should confirm it. The second one expresses that a degree resulting from a combination cannot decrease if the combined degrees increase.

In [5], it has been shown that the syntactic counterpart of the fusion of $\pi_1$ and $\pi_2$ is the following possibilistic base, denoted by $\mathcal{B}_\oplus$ (and sometimes by $B_1 \oplus B_2$) and which is made of the union of:

- the initial bases with new weights defined by:
$$\{(\phi_i, 1-(1-\alpha_i) \oplus 1) : (\phi_i, \alpha_i) \in B_1\} \cup$$
$$\{(\psi_j, 1 - 1 \oplus (1-\beta_j)) : (\psi_j, \beta_j) \in B_2\} \quad (1)$$

- and the knowledge common to $B_1$ and $B_2$ defined by:
$$\{(\phi_i \vee \psi_j, 1-(1-\alpha_i) \oplus (1-\beta_j)) : (\phi_i, \alpha_i) \in B_1\ and\ (\psi_j, \beta_j) \in B_2\}$$

It has been shown that $\pi_{\mathcal{B}_\oplus}(\omega) = \pi_1(\omega) \oplus \pi_2(\omega)$ where $\pi_{\mathcal{B}_\oplus}$ is the possibility distribution associated to $\mathcal{B}_\oplus$ using Definition 1.

In the case of $n$ sources, the syntactic computation of the resulting base can be easily applied when $\oplus$ is associative. Note that it is also possible to provide syntactic counterpart for non-associative fusion operator. In this case $\oplus$ is no longer a binary operator, but a n-ary operator applied to vectors of possibility distributions. The syntactic counterpart is as follows: Let $\mathcal{B} = (B_1, ..., B_n)$ be a vector of possibilistic bases. Let $(\pi_1, \cdots, \pi_n)$ be their associated possibility distributions and $\pi_{\mathcal{B}_\oplus}$ be the result of combining $(\pi_1, ..., \pi_n)$ with $\oplus$. Then, the base associated to $\pi_{\mathcal{B}_\oplus}$ is: $\mathcal{B}_\oplus = \{(D_j, 1 - x_1 \oplus ... \oplus x_n) : j = 1, n\}$, where $D_j$ are disjunctions of size $j$ between formulas taken from different $B_i$'s $(i = 1, n)$ and $x_i$ is either equal to $1-\alpha_i$ or to 1 depending if $\phi_i$ belongs to $D_j$ or not.

## 3 Possibilistic merging operators

This section analyses several classes of $\oplus$ which cope with different issues met in merging multiple sources information. In the rest of this paper, we assume that $\oplus$ is associative.

### 3.1 Conjunctive operators

One of the important aims in merging uncertain information is to exploit complementarities between the sources in order to get a more complete and precise global point of view. Since we deal with prioritized information, two kinds of complementarities can be considered depending on whether we refer to formulas only, or to priorities attached to formulas. In this subsection, we introduce conjunctive operators which exploit the symbolic complementarities between sources.

**Definition 6** $\oplus$ *is said to be a conjunctive operator if* $\forall a \in [0, 1], a \oplus 1 = 1 \oplus a = a$.

The following proposition shows indeed that conjunctive operators, in case of consistent sources of information, exploit their complementarities by recovering all the symbolic information.

**Proposition 1** *Let $B_1$ and $B_2$ be such that $B_1^* \wedge B_2^*$ is consistent. Let $\oplus$ be a conjunctive operator. Then, $\mathcal{B}_\oplus^* \equiv B_1^* \wedge B_2^*$.*

An important feature of a conjunctive operator is its ability to give preference to more specific information. Namely, if an information source $S_1$ contains all the information provided by $S_2$, then combining $S_1$ and $S_2$ with a conjunctive operator leads simply to $S_1$:

**Proposition 2** *Let $B_1$ and $B_2$ be such that $\forall(\psi, \beta) \in B_2, B_1 \vdash_\pi (\psi, \beta)$. Then, $\mathcal{B}_\oplus^* \equiv B_1^*$.*

An example of a conjunctive operator is the *minimum* (for short *min*), for which we can easily check that $\mathcal{B}_\oplus = B_1 \cup B_2$. Other examples are the product, and the geometric average defined by $a \oplus b = \sqrt{ab}$.

### 3.2 Disjunctive operators

Another important issue in fusion information is how to deal with conflicts. When all the sources are equally reliable and conflicting, then one should avoid arbitrary choice by inferring all information provided by one of the sources. Namely, if $B_1 \cup B_2$ is inconsistent, then one can require that $\mathcal{B}_\oplus$ neither infers $B_1$ nor $B_2$. Such a behaviour cannot be captured by any conjunctive operator (See Section 5). This requirement is captured by the disjunctive operators defined by:

**Definition 7** $\oplus$ *is said to be a disjunctive operator if* $\forall a \in [0, 1], a \oplus 1 = 1 \oplus a = 1$.

Then, we have:

**Proposition 3** *Let $B_1$ and $B_2$ be such that $B_1^* \wedge B_2^*$ is inconsistent. Then, there exist $(\phi, \alpha) \in B_1$ and $(\psi, \beta) \in B_2$ such that $\mathcal{B}_\oplus \not\vdash_\pi (\phi, \alpha)$ and $\mathcal{B}_\oplus \not\vdash_\pi (\psi, \beta)$.*

Note that if $\oplus$ is a disjunctive operator then $\mathcal{B}_\oplus$ is of the form: $\mathcal{B}_\oplus = \{(\phi_i \vee \psi_j, 1 - (1 - \alpha_i) \oplus (1 - \beta_j))\}$.
Now, a second natural requirement that one may ask for, in case of conflicts, is to recover the disjunction of all the symbolic information provided by the sources. Clearly, it is easy to find a disjunctive operator which does not satisfy this second requirement. A trivial case is to take the "vacuous" disjunctive operator defined by: $\forall a, \forall b, a \oplus b = 1$.
To satisfy this second requirement we define the notion of regular disjunctive operator:

**Definition 8** *A disjunctive operator $\oplus$ is said to be regular if $\forall a \neq 1, \forall b \neq 1, a \oplus b \neq 1$.*

Then, we have:

**Proposition 4** *Let $B_1$ and $B_2$ be two bases and $\oplus$ be a regular disjunctive operator. Then, $\mathcal{B}_\oplus^* \equiv B_1^* \vee B_2^*$.*

Examples of regular disjunctive operators are the *max*, the so-called *"probabilistic sum"* defined by:
$a \oplus b = a + b - ab$, and the dual of the geometric average defined by $a \oplus b = 1 - \sqrt{(1 - a)(1 - b)}$.
Lastly, note that regular disjunctive operators are not appropriate in the case of consistency between sources; in particular they give preference to less specific information.

### 3.3 Idempotent operators

Another important problem in fusing multiple sources information is how to deal with redundant information. There are two different situations: either we ignore the redundancies, which is suitable when the sources are not independent, or we view redundancy as a confirmation of the same information provided by independent sources. Idempotent operations are defined by:

**Definition 9** $\oplus$ *is said to be an idempotent operator if $\forall a \in [0, 1], a \oplus a = a$.*

Idempotent operators aim to ignore direct redundancies. Namely, if two sources of information entail the same formula $\phi$ to a degree $\alpha$, then one may require that the fused base should not entail $\phi$ with a degree higher than $\alpha$. However, such a requirement is strong since $\phi$ can be obtained from another path exploiting complementarities between higher level formulas provided by the two sources. This is illustrated by the following example:

**Example 1** Let $B_1 = \{(\psi, .9); (\phi, .2)\}$ and
$$B_2 = \{(\phi \vee \neg \psi, .8); (\phi, .2)\}.$$
Clearly $B_1 \vdash_\pi (\phi, .2)$ and $B_2 \vdash_\pi (\phi, .2)$. Now let $\oplus$ be an idempotent operator defined by: $a \oplus b = \frac{a+b}{2}$.
Then, $\mathcal{B}_\oplus = \{(\psi, .45); (\phi \vee \psi, .55); (\phi \vee \neg \psi, .5)\}$ after removing subsumed formulas. We can easily check that $\mathcal{B}_\oplus \vdash_\pi (\phi, .5)$ with $.5 \geq .2$. This is mainly due to the two pieces of information $(\psi, .9)$ and $(\phi \vee \neg \psi, .8)$, provided separately by the sources.

Now, the following proposition shows the cases where idempotent operators indeed ignore redundancies:

**Proposition 5** Let $B_1$ and $B_2$ be two bases, and $\oplus$ be an idempotent operator. Let $\phi$ be such that $B_1 \vdash_\pi (\phi, \alpha)$; $B_2 \vdash_\pi (\phi, \beta)$ with $\beta \leq \alpha$. Let $\Gamma = B_{1 > \alpha} \cup B_{2 > \alpha}$. Then, if $\Gamma \nvdash \phi$ then $\mathcal{B}_\oplus \vdash_\pi (\phi, \gamma)$, with $\gamma \leq max(\alpha, \beta)$.

Note that $\gamma$ may be equal to 0 in case of inconsistency. $\Gamma$ in this proposition is the set of classical formulas in $B_1$ and $B_2$ having a weight strictly greater than $\alpha$. If $\phi$ cannot be deduced from $\Gamma$ then the idempotent property only guarantees that the repeated information will not be inferred with a priority higher than the one with which it can be individually obtained from the different sources.

### 3.4 Reinforcement operators

The aim of reinforcement operators is to view redundancy of information as a confirmation of this information. Namely, if the same piece of information is supported by two different sources, then the priority attached to this piece of information should be strictly greater than the one provided by the sources. A first formal class of reinforcement operators can be defined as follows:

**Definition 10** $\oplus$ is said to be a reinforcement operator if $\forall a, b \neq 1$ and $a, b \neq 0, a \oplus b < min(a, b)$.

We can easily check that if we aggregate the two pieces of information $(\phi, \alpha)$ and $(\phi, \beta)$, then the resulting base is: $\{(\phi, f(\alpha, \beta))\}$ where $f(\alpha, \beta) = 1 - (1-\alpha) \oplus (1-\beta) > max(\alpha, \beta)$ for $\alpha, \beta \in (0, 1)$.
Besides, one can require that reinforcement operations recover all the common information with a higher weight. Namely if the same formula is a plausible consequence of each base, then this formula should be accepted in the fused base with a higher priority. The following proposition shows a first case where this result holds:

**Proposition 6** Let $B_1$ and $B_2$ be such that $B_1^* \wedge B_2^*$ is consistent. Let $\phi$ be such that $B_1 \vdash_\pi (\phi, \alpha)$ and $B_2 \vdash_\pi (\phi, \beta)$ where $\alpha$ and $\beta$ are strictly positive. Let $\oplus$ be a reinforcement operator. Then, $\mathcal{B}_\oplus \vdash_\pi (\phi, \gamma)$, with $\gamma > max(\alpha, \beta)$ if $\alpha, \beta \in (0, 1)$, and $\gamma = 1$ if $\alpha = 1$ or $\beta = 1$.

Now, in case of conflicts, and more precisely, in case of a strong conflict, namely $Inc(B_1 \cup B_2) = 1$, then the above proposition does not hold.
Indeed, let $B_1 = \{(\phi, 1), (\psi, \alpha)\}$ and $B_2 = \{(\neg \phi, 1), (\psi, \beta)\}$. Then we can check that $Inc(\mathcal{B}_\oplus) = 1$, so we cannot infer $\psi$ from $\mathcal{B}_\oplus$ since $Inc(B_1 \cup B_2) = 1$. Even if we add $(\psi, 1)$ to $B_1 \cup B_2$ explicitly then $\psi$ cannot be recovered. In possibilistic logic, when there is a strong conflict then only tautologies are plausible consequences. In this case it is better to use a regular disjunctive operation.
So the first condition is to avoid that $Inc(B_1 \cup B_2) = 1$. But this is not enough since even if $Inc(B_1 \cup B_2) < 1$ one can have $Inc(\mathcal{B}_\oplus) = 1$ due to the reinforcement effect which can push the priority of conflicting information to the maximal priority allowed. For instance let us consider the excessively optimistic reinforcement operator defined by:
$$\forall a, \forall b, a \neq 1, b \neq 1, a \oplus b = b \oplus a = 0.$$
Then we can check that as soon as there is a conflict between the bases to be merged, the inconsistency degree of the fuses base will reach the maximal value.
The following definition focuses on a more interesting class of reinforcement operations:

**Definition 11** A reinforcement operation $\oplus$ is said to be progressive if $\forall a, b \neq 0, a \oplus b \neq 0$.

The progressive operation guarantees that if some formula $(\phi, \alpha)$ with $\alpha > 0$ is inferred by the sources then this formula belongs to $\mathcal{B}_\oplus$ with a weight $\beta$ such that $\alpha < \beta < 1$. However, this new weight $\beta$ can be less than the inconsistency degree of $\mathcal{B}_\oplus$ and therefore $\phi$ will be drowned by the inconsistency of the database. This situation is illustrated by the following example:

**Example 2**
Let $B_1 = \{(\phi \vee \psi, .9); (\phi, .5); (\psi, .5); (\xi, .1)\}$ and $B_2 = \{(\neg \phi \vee \neg \psi, .9); (\neg \phi, .5); (\neg \psi, .5); (\xi, .1)\}$.
Clearly, each base entails $\xi$ which is largely below the inconsistency degree of $B_1 \cup B_2$. Now, let us compute $B_1 \oplus B_2$ with the product operator which is a progressive operator. We get: $B_1 \oplus B_2 = B_1 \cup B_2 \cup \{(\phi \vee \psi \vee \xi, .91); (\neg \phi \vee \neg \psi \vee \xi, .91); (\phi \vee \neg \psi, .75); (\neg \phi \vee \psi, .75); (\phi \vee$ $\xi, .55); (\psi \vee \xi, .55); (\neg\phi \vee \xi, .55); (\neg\psi \vee \xi, .55); (\xi, .19)\}$. Note that there is a reinforcement on $\xi$ since its new weight is .55 (which can be obtained for instance from $(\phi \vee \xi, .55)$ and $(\neg\phi \vee \xi, .55)$). However, this new weight is less than the inconsistency degree of $\mathcal{B}_\oplus$ which is of .75, higher than $Inc(B_1 \cup B_2) = .5$.

The following proposition generalizes Proposition 6, and shows that if the inconsistency degree does not increase, then the common knowledge is entailed.

**Proposition 7** Let $B_1$ and $B_2$ be such that $Inc(B_1 \cup B_2) \neq 1$. Let $\phi$ be such that $B_1 \vdash_\pi (\phi, \alpha)$ and $B_2 \vdash_\pi (\phi, \beta)$ with $\alpha > 0, \beta > 0$. Let $\oplus$ be a progressive reinforcement operation. Then,
if $Inc(\mathcal{B}_\oplus) = Inc(B_1 \cup B_2)$ then, $\mathcal{B}_\oplus \vdash_\pi (\phi, \gamma)$
with $\gamma > max(\alpha, \beta)$, and $\gamma = 1$ if $\alpha = 1$ or $\beta = 1$.

### 3.5 Adaptive merging operators

The regular disjunctive operators appear to be appropriate when the sources are completely conflicting. However, in the case of consistency, or of a low level of inconsistency regular disjunctive operators are very cautious. Besides, reinforcement is not appropriate in the case of complete conflicts.
The aim of adaptive operators is to have a disjunctive behaviour in a case of complete contradiction and the progressive reinforcement behaviour in the other case. Let $\oplus_d$ and $\oplus_r$ be respectively a regular disjunctive and progressive reinforcement operators. Let $h$ be either equal to 1 or to 0. Then we define an adaptive operation, denoted by $\oplus_h$, as follows:
$a \oplus_h b = max(min(h, (a \oplus_d b)), min(1 - h, (a \oplus_r b)))$.
Then we have the following result:

**Proposition 8** Let $B_1$ and $B_2$ be two possibilistic bases. Let $h$ be equal to 1 if $Inc(B_1 \cup B_2) = 1$ and equal to 0 otherwise. Let $\oplus_h$ be an adaptive operator. If $Inc(\mathcal{B}_\oplus) = Inc(B_1 \cup B_2)$ then, $\forall \phi$, if $B_1 \vdash_\pi (\phi, \alpha)$ and $B_2 \vdash_\pi (\phi, \beta)$ then we have: $\mathcal{B}_{\oplus_h} \vdash (\phi, \gamma)$ with $\gamma > 0$.

### 3.6 Averaging operators

A last class of merging operators which is worth considering is the so-called *averaging operation*, well known for aggregating preferences, and defined by:

**Definition 12** $\oplus$ is called an averaging operator if
$max(a, b) \geq a \oplus b \geq min(a, b),$
with $\oplus \neq max$ and $\oplus \neq min$.

One example of averaging operators is the arithmetic mean $a \oplus b = \frac{a+b}{2}$. In this case, at the syntactic level, the result of combining $B_1$ and $B_2$ writes:
$\{(\phi_i, \frac{\alpha_i}{2})\} \cup \{(\psi_j, \frac{\beta_j}{2})\} \cup \{(\phi_i \vee \psi_j, \frac{\alpha_i + \beta_j}{2})\}$.
From this writing, in case of consistency we can check that if $B_1 \vdash_\pi (\phi, \alpha)$ and $B_2 \vdash_\pi (\phi, \beta)$ then $\mathcal{B}_\oplus \vdash_\pi (\phi, \gamma)$ with $\gamma \geq \frac{\alpha+\beta}{2}$.

### 3.7 Accounting for reliabilities of the sources

The possibilistic logic framework enables us to take also into account priorities between sources (or agents). Here priority may mean either that the sources are decreasingly ordered according to their reliability, or that a reliability degree is attached to each source. When we have just a reliability ordering and no commensurability assumption is made between the scales used for stratifying each source, the approach which can be used is known in social choice theory under the name of "dictatorship". The idea is to refine one ranking by the other. More precisely, let $\pi_1$ and $\pi_2$ be two possibility distributions. Assume that $\pi_1$ has priority over $\pi_2$. The result of combination defined by:
*i.* If $\pi_1(\omega) > \pi_1(\omega')$ then $\pi_\oplus(\omega) > \pi_\oplus(\omega')$
*ii.* If $\pi_1(\omega) = \pi_1(\omega')$ then $\pi_\oplus(\omega) \geq \pi_\oplus(\omega')$ iff $\pi_2(\omega) \geq \pi_2(\omega')$.
Clearly the combination result is simply the refinement of $\pi_1$ (the dictator) by $\pi_2$. Syntactic counterpart of this combination can be found in [5].
When a reliability degree is associated with each source, we may use weighted counterparts of operations $\oplus$. However in practice, it amounts to performing a preliminary modification of the degrees attached to formulas provided by each source and then to performing a non-weighted combination operation on the modified possibilistic bases. For instance, using the weighted min conjunction defined by $\forall \omega, \pi_\oplus(\omega) = min_{j=1,n} max(\pi_j(\omega), 1 - \lambda_j)$ (for $\lambda_j = 1, \forall j$, the min combination is recovered). It amounts to performing the union of discounted bases of the form $Discount(B_i, \lambda_i) = \{(\phi, \lambda_i) | (\phi, \beta) \in B_i \text{ and } \beta \geq \lambda_i\} \cup \{(\phi, \beta) | (\phi, \beta) \in B_i \text{ and } \beta < \lambda_i\}$. It is worth pointing out that discounting sources help solve conflicts between sources in a natural way.

## 4 Postulates for classical merging

Let us first introduce some additional notations.
Let $E = \{K_1, ..., K_n\}$ $(n \geq 1)$ be a multi-set of $n$ propositional bases to be merged. $E$ is called an information set. $\wedge E$ (resp. $\vee E$) denotes the conjunction (resp. disjunction) of the propositional bases of $E$. The symbol $\sqcup$ denotes the union on multi-sets.
For the sake of simplicity, if $K$ and $K'$ are propositional bases and $E$ an information set we simply write $E \sqcup K$ and $K \sqcup K'$ instead of $E \sqcup \{K\}$ and $\{K\} \sqcup \{K'\}$ respectively. We will denote $K^n$ the multi-set $\{K, ..., K\}$ of size $n$. A classical merging operator $\Delta$ is a function applied on $E$ and which returns a classical base, denoted by $\Delta(E)$.
Koniesczny and Pino Pérez [10] have proposed a set of basic properties that a merging operator has to satisfy:
$(A_1)$ $\Delta(E)$ is consistent;
$(A_2)$ If $E$ is consistent, then $\Delta(E) = \wedge E$;
$(A_3)$ If $E_1 \leftrightarrow E_2$, then $\vdash \Delta(E_1) \equiv \Delta(E_2)$;
$(A_4)$ If $K \wedge K'$ is inconsistent, then $\Delta(K \sqcup K') \not\vdash K$;
$(A_5)$ $\Delta(E_1) \wedge \Delta(E_2) \vdash \Delta(E_1 \sqcup E_2)$;
$(A_6)$ If $\Delta(E_1) \wedge \Delta(E_2)$ is consistent, then
$\Delta(E_1 \sqcup E_2) \vdash \Delta(E_1) \wedge \Delta(E_2)$;
where $E_1 \leftrightarrow E_2$ means that there exists a bijection $f$ from $E_1 = \{K_1^1, ..., K_n^1\}$ to $E_2 = \{K_1^2, ..., K_n^2\}$ such that $\forall K \in E_1, \exists K' \in E_2, f(K) \equiv K'$.
Liberatore [12], in the context of commutative belief revision, does not impose $A_1$. He allows the result to be inconsistent if the bases to merge are individually inconsistent. Moreover, he gives another postulate in the same spirit as $A_4$, namely:
$(A_7)$ $\Delta(K \sqcup K') \vdash K \vee K'$.
Two classes of merging operators have been particulary analysed in the literature: majority operators defined by: $(Maj)$ $\forall K, \exists n, \Delta(E \sqcup K^n) \vdash K$,
and arbitration operators defined by:
$(Arb)$ $\forall K, \forall n, \Delta(E \sqcup K^n) = \Delta(E \sqcup K)$.

## 5 Postulates for possibilistic merging

This section relates the general classes of possibilistic merging operations to the rational postulates recalled in the previous section. But first we need an adaptation of these postulates in order to take into account the priorities attached to the information. A first immediate way of adapting the classical postulates is to require that the result of the merging be a classical base. If $\mathcal{B}_\oplus$ denotes the result of merging $\mathcal{B} = \{B_1, ..., B_n\}$ with $\oplus$, then the classical base resulting from merging $B_i$'s with $\oplus$ is simply:
$$\Delta_\oplus(\mathcal{B}) = \{\phi : (\phi, \alpha) \in \mathcal{B}_\oplus, \alpha > Inc(\mathcal{B}_\oplus)\}.$$
However, restricting the result of merging prioritized bases to a classical base is not satisfactory. Indeed it leads to lose the associativity property of associative operators. The natural question is how to define $\Delta_\oplus(B_1, \Delta_\oplus(B_2, B_3))$ since $B_1$ is a stratified base, while $\Delta_\oplus(B_2, B_3)$ is a classical one. One way of enforcing the iteration is to give to formulas of the resulting classical base a weight equal to 1. However, this may violate the reliability of the formulas since formulas in $\Delta_\oplus(B_1, B_2)$ which were very uncertain become fully reliable. The loss of associativity property is illustrated by the following example:

**Example 3** Let $\mathcal{B} = \{B_1, B_2, B_3\}$ such that $B_1 = \{(\phi, .8)\}$, $B_2 = \{(\neg\phi, .5); (\psi, .4)\}$ and $B_3 = \{(\psi, .3)\}$. Let $\oplus = min$. We have $\Delta_\oplus(B_1, B_2) = \{\phi\}$. To be able to merge this result with $B_3$ we associate to $\phi$ a weight equal to 1, and we get $\Delta_\oplus(\Delta_\oplus(B_1, B_2), B_3) = \{\phi, \psi\}$. We also have $\Delta_\oplus(B_2, B_3) = \{\neg\phi, \psi\}$ and $\Delta_\oplus(B_1, \Delta_\oplus(B_2, B_3)) = \{\neg\phi, \psi\}$. Then, $\Delta_\oplus(\Delta_\oplus(B_1, B_2), B_3) \not\equiv \Delta_\oplus(B_1, \Delta_\oplus(B_2, B_3))$.

### 5.1 Adapting classical postulates

We focus on the approach where the result of the merging operation is a stratified base. Therefore, the process of merging can be iterated. Let us now adapt the classical postulates recalled in Section 4.

Let us adapt $(A_1)$. Possibilistic logic, contrary to classical logic, does not infer anything in the presence of inconsistency. Hence a partially inconsistent base $\mathcal{B}_\oplus$ (with $Inc(\mathcal{B}_\oplus) < 1$) can be still meaningful, since plausible conclusions can be inferred from it, by taking its consistent part, i.e. the set of formulas having a weight greater than $Inc(\mathcal{B}_\oplus)$. Thus, the adaptation of $(A_1)$ can be weakened as follows:
$(P_1)$ $\mathcal{B}_\oplus$ is not fully inconsistent, i.e., $Inc(\mathcal{B}_\oplus) < 1$.
Note that if one insists on providing a consistent and stratified base as a result of fusion, then the associativity can be lost for associative operations.

Let us adapt the second postulate $(A_2)$. Requiring an equivalence between $\mathcal{B}_\oplus$ and $B_1 \cup \cdots \cup B_n$ in the second postulate is very strong with stratified bases. For instance, assume that two identical formulas $(\phi, \alpha)$ have to be aggregated. We have already seen that with a reinforcement operator we get $(\phi, \beta)$ $(\beta > \alpha)$ as a result of the merging. So, we do not recover the initial weight of $\phi$. We propose to weaken $A_2$ as follow:
$(P_2)$ If $B_1 \cup \cdots \cup B_n$ is consistent, then $\mathcal{B}_\oplus \vdash_\pi (\phi, \beta)$ iff $B_1 \cup \cdots \cup B_n \vdash_\pi (\phi, \gamma)$, with $\beta > 0$ and $\gamma > 0$.
This postulate implies that if $B_1^* \wedge ... \wedge B_n^*$ is consistent, then $\mathcal{B}_\oplus^* \equiv B_1^* \wedge \cdots \wedge B_n^*$.

Postulates $A_3$ and $A_4$ have immediate counterparts:
Let $\mathcal{B} = \{B_1, \cdots, B_n\}$ and $\mathcal{B}' = \{B_1', \cdots, B_n'\}$.
$(P_3)$ If $\mathcal{B} \leftrightarrow \mathcal{B}'$, then $\mathcal{B}_\oplus \equiv_s \mathcal{B}'_\oplus$,
where $\mathcal{B} \leftrightarrow \mathcal{B}'$ means that there exists a bijection $f$ from $\mathcal{B}$ to $\mathcal{B}'$ such that $\forall B \in \mathcal{B}, \exists B' \in \mathcal{B}', f(B) \equiv_s B'$.
$(P_4)$ If $B_1 \cup B_2$ is inconsistent, then $\mathcal{B}_\oplus \not\vdash_\pi B_1$ and $\mathcal{B}_\oplus \not\vdash_\pi B_2$.
Concerning postulates $A_5$ and $A_6$, notice that in classical logic, when $\Delta(E_1) \wedge \Delta(E_2)$ is inconsistent then $A_5$ is trivially satisfied. Hence $A_5$ is only meaningful when there is no conflict between the sources. Therefore $A_5$ and $A_6$ are adapted as follows:
$(P_5)$ If $\mathcal{B}_\oplus$ is consistent with $\mathcal{B}'_\oplus$, then
$$\mathcal{B}_\oplus \cup \mathcal{B}'_\oplus \vdash_\pi (\mathcal{B} \cup \mathcal{B}')_\oplus.$$
$(P_6)$ If $\mathcal{B}_\oplus$ is consistent with $\mathcal{B}'_\oplus$, then
$$(\mathcal{B} \cup \mathcal{B}')_\oplus \vdash_\pi \mathcal{B}_\oplus \cup \mathcal{B}'_\oplus.$$
Let us now see how to adapt the postulate $A_7$.
The common knowledge in the prioritized case can be defined as follows: If $B_1 \vdash_\pi (\phi, \alpha)$ and $B_2 \vdash_\pi (\phi, \beta)$, then $\mathcal{B}_\oplus \vdash_\pi (\phi, \gamma)$, with $\gamma > 0$.
Now, the question is how to fix the value of $\gamma$. Ob-

viously $\gamma$ should not be greater than $min(\alpha,\beta)$. Indeed assume that $c > a \geq b$, and $B_1 = \{(\phi,\alpha)\}$ and $B_2 = \{(\phi,\beta)\}$. Then it can be checked that $(\phi,\gamma)$ is not a consequence of $B_2$, which means that $(\phi,\gamma)$ cannot be considered as a common information of $B_1$ and $B_2$. Therefore the adaptation of $A_7$ is as follows:
$(P_7)$ $\forall \phi$, if $B_1 \vdash_\pi (\phi,\alpha)$ and $B_2 \vdash_\pi (\phi,\beta)$ then $\mathcal{B}_\oplus \vdash_\pi (\phi,\gamma)$ with $0 < \gamma \leq min(\alpha,\beta)$.

Lastly, arbitration and majority have immediate extensions: $(Arb)$ $\forall B, \forall n, (\mathcal{B} \cup B^n)_\oplus \equiv_s (\mathcal{B} \cup B)_\oplus$, and $(Maj)$ $\forall B, \exists n, (\mathcal{B} \cup B^n)_\oplus \vdash_\pi B$.

## 5.2 Properties of the fusion operations

This section gives the properties of the classes of possibilistic operators introduced in Section 3.
Proposition 9 shows that $\oplus$ is syntax independent.

**Proposition 9** *Any possibilistic merging operation satisfies $P_3$.*

The next proposition relates the property of idempotency to the idea of arbitration:

**Proposition 10** *Any idempotent operation is an arbitration operation.*

The following proposition gives the properties of the regular disjunctive operations.

**Proposition 11** *Let $\oplus$ be a regular disjunctive operator. Then, $\oplus$ satisfies $P_1, P_4, P_5, P_7$ but may fail to satisfy $P_2, P_6, Maj, Arb$.*

**Counter-examples**: For $P_2, P_6$ and $Maj$ we use the operator $\oplus = max$ which is a regular disjunctive operation:

- $P_2$, $P_6$: Let $\mathcal{B}= \{B_1, B_2\}$ with $B_1 = \{(\phi,.8)\}$ and $B_2 = \{(\psi,.3)\}$. Although $B_1 \cup B_2$ is consistent, we have $\mathcal{B}_\oplus = \{(\phi \lor \psi, .3)\}$ and we recover neither $B_1^*$ nor $B_2^*$. Then, $\oplus$ does not satisfy $P_2$. It does not satisfy $P_6$ for the same reason.

- $Maj$: $max$ is an idempotent operator, hence it is an arbitration operation, and cannot be a majority operator.

For $Arb$, consider the probabilistic sum defined by: $a \oplus b = a + b - ab$.
Let $B = \{(\phi,\alpha)\}$. Then, one can easily check that $B \oplus B = \{(\phi, 2\alpha - \alpha^2)\}$ which is different from $B = \{(\phi,\alpha)\}$.

**Proposition 12** *Let $\oplus$ be a conjunctive operator. Then, $\oplus$ satisfies $P_2, P_6$ but may fail to satisfy $P_4, P_5, P_7, Arb, Maj$.*

**Counter-examples:**

- For $P_4$ and $P_7$, let us use the $min$ since it is a conjunctive operator.

  Let $B_1 = \{(\neg\phi,.6); (\psi,.5)\}$ and $B_2 = \{(\phi,.7)\}$.
  Since $Inc(\mathcal{B}_\oplus) = .6$, the useful information of $\mathcal{B}_\oplus$ is $\{(\phi,.7)\} = B_2$. Then, $P_4$ is not satisfied.
  Let now $B_2 = \{(\phi,.7); (\psi,.5)\}$. Then, $(\psi,.5)$ cannot be inferred from $\mathcal{B}_\oplus$, and $P_7$ is not satisfied.
  $min$ is idempotent, then it cannot be a majority operator.

- For $P_5$ and $Arb$, let us consider the product which is a conjunctive operator.
  Let $\mathcal{B}= \{B_1\}$ and $\mathcal{B}' = \{B_2\}$ s.t. $B_1 = \{(\phi,.5)\}$ and $B_2 = \{(\psi,.6)\}$. We have $\mathcal{B}_\oplus \cup \mathcal{B}'_\oplus = \{(\phi,.5); (\psi,.6)\}$ and $(\mathcal{B} \cup \mathcal{B}')_\oplus = \{(\phi,.5); (\psi,.6); (\phi \lor \psi, .8)\}$. $(\phi \lor \psi, .8)$ cannot be inferred from $\mathcal{B}_\oplus \cup \mathcal{B}'_\oplus$. Then, $P_5$ is not satisfied. We also have $B_1 \oplus B_2 = \{(\phi,.5); (\psi,.6); (\phi \lor \psi, .8)\}$ and $B_1 \oplus B_2^2 = \{(\phi,.5); (\psi,.84); (\phi \lor \psi, .92)\}$, then $Arb$ is not satisfied.

The following proposition relates the property of majority to the reinforcement property.

**Proposition 13** *Let $B_1$ be a possibilistic base, and $B_2$ another possibilistic base which is not conflicting with completely certain formulas of $B_1$. Let $\oplus$ be a progressive reinforcement operator. Denote by $B_2^n$ the combination of $B_2$ n times with $\oplus$. Then, $\exists n, \forall (\psi,\beta) \in B_2$, $B_1 \oplus B_2^n \vdash_\pi (\psi,\gamma)$ with $\gamma > \beta$, ($\gamma = 1$ if $\beta = 1$).*

This proposition means that reinforcement operators are majority operators, in the sense that if the same piece of information is repeated enough times then this piece of information will be believed.
This proposition does not hold if we only use reinforcement operations which are not progressive. For instance, consider the *Luckasiewicz t-norm* defined by: $a \oplus b = max(0, a+b-1)$.
Then, for instance consider the bases $B_1 = \{(\phi,.8)\}$, $B_2 = \{(\phi,.8)\}$ and $B = \{(\neg\phi,.7)\}$ which are not completely conflicting. Then, we can easily check that $Inc(B_1 \oplus B_2 \oplus B^2) = 1$ and hence $B$ cannot be deduced. Indeed, we have $B_1 \oplus B_2 = \{(\phi,1)\}$, $B^2 = B \oplus B = \{(\neg\phi, 1)\}$.
We now give the properties of progressive reinforcement operators:

**Proposition 14** *Let $\oplus$ be a progressive reinforcement operator. Then, $\oplus$ satisfies $P_1$ (provided that $Inc(B_1 \cup \cdots \cup B_n) < 1$), $P_2, P_6, P_7$ (provided that $Inc(\mathcal{B}_\oplus) = Inc(B_1 \cup \cdots \cup B_n)$ and $Inc(B_1 \cup \cdots \cup B_n) < 1$), $Maj$ but may fail to satisfy $P_4, P_5, Arb$.*

**Counter-example:** Let us use the product which is a progressive reinforcement operator.

- $P_4$: Let $B_1 = \{(\phi,.6)\}$ and $B_2 = \{(\neg\psi,.5)\}$. The useful information (above the level of inconsistency) of $\mathcal{B}_\oplus$ is $\{(\phi,.6)\} = B_1$. Then, $P_4$ is not satisfied.

- $P_5, Arb$: see the counter-example of Proposition 12.

The following proposition summarizes the properties of averaging operators:

**Proposition 15** *The Averaging operator satisfies $P_2, P_4, P_5, P_6$ and Arb but may fail to satisfy $P_7$ and Maj.*

Lastly, the following tree summarizes the considered operators, with the associated satisfied postulates:

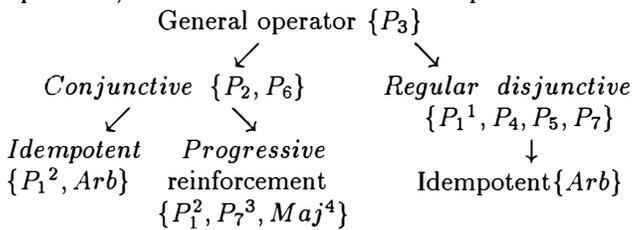

General operator $\{P_3\}$
- Conjunctive $\{P_2, P_6\}$
  - Idempotent $\{P_1^2, Arb\}$
  - Progressive reinforcement $\{P_1^2, P_7^3, Maj^4\}$
- Regular disjunctive $\{P_1^1, P_4, P_5, P_7\}$
  - Idempotent $\{Arb\}$

Note that there exists conjunctive operators which do not satisfy $P_1$ like *Luckasiewicz t-norm*.

In the following table, we consider the three noticeable possibilistic operators $min, max$ and the product $Pro$. The symbol $\sqrt{}$ (resp. $-$) means that the operator satisfies (resp. falsify) the postulate.

|     | $P_1$ | $P_2$ | $P_3$ | $P_4$ | $P_5$ | $P_6$ | $P_7$ | Arb | Maj |
|-----|-------|-------|-------|-------|-------|-------|-------|-----|-----|
| min | $\sqrt{}^2$ | $\sqrt{}$ | $\sqrt{}$ | $-$ | $\sqrt{}$ | $\sqrt{}$ | $-$ | $\sqrt{}$ | $-$ |
| max | $\sqrt{}$ | $-$ | $\sqrt{}$ | $\sqrt{}$ | $\sqrt{}$ | $-$ | $\sqrt{}$ | $\sqrt{}$ | $-$ |
| Pro | $\sqrt{}^2$ | $\sqrt{}$ | $\sqrt{}$ | $-$ | $-$ | $\sqrt{}$ | $\sqrt{}^3$ | $-$ | $\sqrt{}^4$ |

**Table 1**

In [3] we have shown that $min$ and $Pro$ operators are the possibilistic counterparts of $max$ and $\Sigma$ respectively, proposed in the classical merging [10, 11]. By comparing the above table with the one presented in [10] for classical merging operators we see that $P_4$ is satisfied by $max$ and $\Sigma$ operators but it is not by $min$ and $Pro$. This is due to the presence of priorities in the possibilistic framework. Then in the presence of inconsistency, we may favor a base if its formulas are more reliable. Moreover, when the formulas are weighted we can express the reinforcement effect which explains that $P_5$ is not satisfied by $Pro$.

## 6 Conclusion

Possibilistic logic acknowledges the presence of a stratification between classical logic formulas in the inference process. This stratification which reflects certainty degrees or priorities is particularly useful for dealing with conflicts in the fusion process (even approaches to fusion in the classical logic setting use implicit stratifications based on Dalal distance [7]). The logical setting is well suited in practice for expressing knowledge or preferences in a granular and high level way. Thus the typology of the fusion operations in the possibilistic setting provides a basis for designing fusion systems able to propose a synthesis of partially conflicting goals on the basis of some chosen type of combination, possibly taking into account priorities between agents or sources. Lastly, this paper has analysed the possibilistic merging operators in terms of postulates. A message from Table 1 is that some postulates which make sense in classical fusion, like $A_4$, are not appropriate for merging prioritized bases. Clearly, future work is to study new postulates proper for prioritized bases.

---

[1] even if $Inc(B_1 \cup \cdots \cup B_n) = 1$.

[2] if $Inc(B_1 \cup \cdots \cup B_n) < 1$.

[3] if $Inc(B_\oplus) = Inc(B_1 \cup B_2)$ and $Inc(B_1 \cup B_2) < 1$.

[4] if $B$ does not contradict completely certain formulas.